\documentclass[10pt,a4paper,conference]{IEEEtran}
\usepackage[latin9]{inputenc}
\usepackage{color}
\usepackage{booktabs}
\usepackage{mathtools}
\usepackage{graphicx}
\makeatletter

\ifCLASSINFOpdf
\else
\fi
\usepackage{caption}
\usepackage{url}
\usepackage{amsfonts}

\@ifundefined{showcaptionsetup}{}{%
 \PassOptionsToPackage{caption=false}{subfig}}
\usepackage{subfig}
\usepackage{authblk}

\makeatother

\begin{document}

\title{Action Classification via Concepts and Attributes}

\author[1]{Amir Rosenfeld}
\author[2]{Shimon Ullman}
\affil[1]{Department of Electrical Engineering and Computer Science\protect\\York University, Toronto, ON, Canada}
\affil[2]{Department of Computer Science \& Applied Mathematics\protect\\Weizmann Institute of Science, Rehovot, Israel}
\affil[1]{\texttt {amir@eecs.yorku.ca}}
\affil[2]{\texttt {shimon.ullman@weizmann.ac.il}}

\maketitle
\begin{abstract}
Classes in natural images tend to follow long tail distributions.
This is problematic when there are insufficient training examples
for rare classes. This effect is emphasized in compound classes, involving
the conjunction of several concepts, such as those appearing in action-recognition
datasets. In this paper, we propose to address this issue by learning
how to utilize common visual concepts which are readily available.
We detect the presence of prominent concepts in images and use them
to infer the target labels instead of using visual features directly,
combining tools from vision and natural-language processing. We validate
our method on the recently introduced HICO dataset reaching a mAP
of 31.54\% and on the Stanford-40 Actions dataset, where the proposed
method outperforms that obtained by direct visual features, obtaining
an accuracy 83.12\%. Moreover, the method provides for each class
a semantically meaningful list of keywords and relevant image regions
relating it to its constituent concepts.
\end{abstract}

\section{Introduction\protect\footnote{This work was done in the Weizmann Institute.}}

In many tasks in pattern recognition, and specifically in computer
vision, target classes follow a long-tail distribution. In the domain
of action recognition this particularly true since the product of
actions and objects is much bigger than each alone, and some examples
may not be observed at all. This has been observed in several studies
\cite{ramanathan2015learning,zhu2014capturing,salakhutdinov2011learning}
and it is becoming increasingly popular to overcome this problem by
building ever larger datasets \cite{lin2014microsoft,ILSVRC15}. However,
the distribution of these datasets will inevitably be long-tailed
as well. One way to tackle this problem is to borrow information from
external data sources. For instance, it has become popular to combine
language and vision using joint embedded spaces \cite{norouzi2013zero,li2015zero},
which allow recognizing unseen classes more reliably than using a
purely visual approach.

In this work, we propose to use an annotated \emph{concept }dataset
to learn a mapping from images to concepts with a visual meaning (e.g.,
objects and object attributes). This mapping is then used as a feature
representation for classifying a \emph{target} dataset, instead of
describing the images with visual features extracted directly. This
allows to describe an image or scene directly with high-level concepts
instead of using visual features directly for the task. We show that
training image classifiers this way, specifically in action-recognition,
is as effective as training on the visual features. Moreover, we show
that the concepts learned to be relevant to each category carry semantic
meaning, which enables us to gain further insights into their success
and failure modes. Our concept dataset is the Visual-Genome dataset
\cite{journals/corr/KrishnaZGJHKCKL16}, in which we leverage the
rich region annotations.

\section{Previous Work}

We list several lines of word related to ours. An early related work
is ObjectBank \cite{li2010object}, where the outputs of detectors
for 200 common objects are aggregated via a spatial-pyramid to serve
as feature representations. In the same spirit, ActionBank \cite{sadanand2012action}
learns detectors for various action types in videos and uses them
to represent others as weighted combinations of actions. The work
of \cite{lampert2009learning} learns object attributes to describe
objects in a zero-shot learning setting, so that new classes (animals)
can be correctly classified by matching them to human generated lists
of attributes. Recently, \cite{ramanathan2015learning} learned various
types of relations between actions (e.g., part of / type of / mutually
exclusive) via visual and linguistic cues and leveraged those to be
able to retrieve images of actions from a very large variety (27k)
of action descriptions. Recent works have shown that emergent representations in deep networks
tend to align with semantic concepts, such as textures, object parts
and entire objects \cite{bau2017network}, confirming
that it may prove useful to further guide a classifier of a certain
class with semantically related concepts. Other works leverage information from natural language: in \cite{norouzi2013zero} an image is mapped to a semantic embedding space by a convex combination
of word embeddings according to a pre-trained classifier on ImageNet
\cite{ILSVRC15}, allowing to describe unseen classes as combinations
of known ones. \cite{li2015zero} makes this more robust by considering
the output of the classifier along with the WordNet \cite{miller1995wordnet}
hierarchy, generating image tags more reliably. The work of \cite{gao2016acd}
mines a large image-sentence corpora for actor-verb-object triplets
and clusters them into groups of semantically related actions. Recently,
\cite{mallya2016learning} used detected or provided person-bounding
boxes in a multiple-instance learning framework, fusing global image
context and person appearance. \cite{gao2015deep} uses spatial-pyramid
pooling on activations from a convolutional layer in a network and
encodes them using Fisher Vectors \cite{gao2015deep}, with impressive
results. 

In our work we do not aim for a concept dataset with only very common
objects or one that is tailored to our specific target task (such
as action-recognition or animal classification) and automatically
learn how to associate the learned concepts to target classes, either
directly or via a language-based model. 

\section{Approach\label{sec:Approach}}

We begin by describing our approach at a high level, and elaborate
on the detail in subsequent sub-sections. 

Our goal is to learn a classifier for a given set of images and target
labels. Assume we are given access to two datasets: (1) a target\emph{
}dataset\emph{ }$\mathcal{F}$ and (2) a concept\emph{ }dataset $\mathcal{D}$.
The target dataset contains training pairs $(I_{i},y_{i})$ of images
$I_{i}$ and target labels $y_{i}$. The concept dataset $\mathcal{D}$
is an additional annotated dataset containing many images labeled
with a broad range of common concepts. The general idea is to learn
high-level concepts from the dataset $\mathcal{D}$ and use those
concepts to describe the images in $\mathcal{F}$. More formally:
let $C=(c_{1},c_{2}\dots c_{N})$ be a set of $N$ concepts appearing
in $\mathcal{D}$. We learn a set of concept classifiers $F_{c}$,
one for each $c\in C$. Once we have the concept classifiers, we use
them to describe each image $I\in\mathcal{F}$ : we apply each classifier
$F_{c}$ to the image $I,$ obtaining a set of concept scores: 
\begin{equation}
S(I)\coloneqq[F_{1}(I),F_{2}(I),\dots F_{N}(I)]\in\mathbb{R}{}^{N}\label{eq:concept-scores}
\end{equation}

For brevity, we'll use the notation $S_{I}=S(I)$. $S_{I}$ defines
a mapping from the samples $I$ to a concept-space. We use $S_{I}$
as a new feature-map, allowing us to learn a classifier in terms of
concepts, instead of features extracted directly from each image $I.$
We note that the dataset $\mathcal{D}$ from which we learn concepts
should be rich enough to enable learning of a broad range of concepts,
to allow to describe each image in $\mathcal{F}$ well enough to facilitate
the classification into the target labels. 

Next, we describe the source of our concept space, and how we learn
various concepts.

\subsection{Learning Concepts}

To learn a broad enough range of concepts, we use the recently introduced
Visual Genome (VG) dataset \cite{journals/corr/KrishnaZGJHKCKL16}.
It contains 108,249 images, all richly annotated with bounding boxes
of various regions within each image. Each region spans an entire
objects or an object part, and is annotated with a noun, an attribute,
and a natural language description, all collected using crowd-sourcing
(see \cite{journals/corr/KrishnaZGJHKCKL16} for full details). The
average number of objects per image is 21.26, for a total of 2.1 million
object instances. In addition, it contains object pair relationships
for a subset of the object pairs of each image. Its richness and diversity
makes it an ideal candidate from which to learn various concepts.

\subsection{Concepts as Classifiers\label{subsec:Concepts-as-Classifiers}}
\begin{center}
\begin{table}
\begin{centering}
\begin{tabular}{ll}
\toprule 
Class  & Top assigned keywords\tabularnewline
\midrule 
brushing\_teeth  & toothbrush, sink, bathroom, rug, brush \tabularnewline
cutting\_trees  & bark, limb, tree\_branch, branches, branch \tabularnewline
fishing  & shore, mast, ripple, water, dock \tabularnewline
holding\_an\_umbrella  & umbrella, rain, handbag, parasol, raincoat \tabularnewline
phoning  & cellphone, day, structure, bathroom, square \tabularnewline
pushing\_a\_cart  & cart, crate, boxes, luggage, trolley \tabularnewline
rowing\_a\_boat  & paddle, oar, raft, canoe, motor \tabularnewline
taking\_photos  & camera, cellphone, phone, lens, picture \tabularnewline
walking\_the\_dog  & dog, leash, tongue, paw, collar \tabularnewline
writing\_on\_a\_board  & writing, racket, poster, letter, mask \tabularnewline
\bottomrule
\end{tabular}
\par\end{centering}
\caption{\label{tab:Highest-ranking-concepts}Highest ranking concepts linked
to each action class according to the proposed method, for 10 arbitrarily
selected actions from the Stanford-40 Actions dataset\cite{yao2011human}.
We train classifiers to detect actions by a weighted sum of detected
image concepts. Most detected keywords are semantically meaningful
(holding\_an\_umbrella$\rightarrow$rain) while some point to dataset
bias (holding\_an\_umbrella$\rightarrow$handbag}
\end{table}
\par\end{center}

We explore the use of three sources of information from the VG dataset,
namely (1) object annotations (2) attribute annotations (group all
objects with a given attribute to a single entity) and (3) object-attributes
(a specific object with a specific attribute). We assign each image
a binary concept-vector $P\in\mathcal{\mathbb{R}}{}^{N}$ where $P_{i}$
indicates if the $i$'th concept is present or not in the respective
image. This is done separately each of the above sources of information.
Despite the objects being annotated via bounding boxes, rather than
training detectors, we train image-level predictors for each. This
is both simpler and more robust, as it can be validated from various
benchmarks (\cite{everingham2010pascal,ILSVRC15} and many others)
that detecting the presence of objects in images currently works more
accurately than correctly localizing them. Moreover, weakly-supervised
localization methods are becoming increasingly effective \cite{bilen2015weakly,cinbis2015weakly,pathak2015constrained},
further justified the use of image-level labels. Given labellings
for $N$ different concepts (where $N$ may vary depending on the
type of concept), we train a one-versus-all SVM for each one separately,
using features extracted from a CNN (see experiments for details).
Denote these classifiers as $(F_{i})_{i=1}^{N}$.

This process results in a scoring of each image $I\in\mathcal{F}$
(our target dataset) with a concept-feature $S_{I}$ as in Eqn. \ref{eq:concept-scores}.
For each concept score $S_{I,j}=F_{j}(I)$, we have 

\begin{eqnarray}
S_{I,j} & = & \upsilon_{j}^{T}f_{I}\\
S_{I} & = & Vf_{I}
\end{eqnarray}

where $f_{I}$ are features extracted from image $I$ and $\upsilon_{j}$
the weight vector of the $j$'th classifier (we drop the bias term
for brevity). $V$ is a matrix whose rows are the $v_{j}$. 

We then train a classifier to map from $S_{I}$ to its target label,
by using an additional SVM for each target class; denote by $(G_{j})_{j=1}^{L}$
each classifier, where $L$ is the number target labels of images
in target dataset $\mathcal{F}$. The final score assigned to an image
$I$ for a target class $j$ is denoted by 

\begin{eqnarray}
H_{j}(I) & = & \omega_{j}^{T}S_{I}\\
 & = & \omega_{j}^{T}Vf_{I}
\end{eqnarray}

Where $\omega_{j}$ is the learned weight vector for the classifier
$G_{j}.$ Before we train $G_{i}$ , we apply PCA to the collection
of training $S_{I}$ vectors and project them to first $n=900$ dimensions
according to the strongest singular values ($n$ was chosen by validation
in early experiments). We found this to reduce the runtime and improve
the classification results.

We also experimented with training a neural net to predict the class
scores; this brought no performance gain over the SVM , despite trying
various architectures and learning rates. 

We next describe how we deal with training a large number of concept
classifiers using the information from the VG dataset.

\subsection{Refinement via Language \label{subsec:Refinement-via-Language}}

The object nouns and attributes themselves have very long-tail distributions
in the VG dataset, and contain many redundancies. Training classifiers
for all of the concepts would be unlikely for several reasons: first,
the objects themselves follow a long-tail distribution, which would
cause most of the classifiers to perform poorly; second, there are
many overlapping concepts, which cannot be regarded as mutually exclusive
classes; third, the number of parameters required to train such a
number of classifiers would become prohibitively large. 

To reduce the number of concepts to be learned, we remove redundancies
by standardizing the words describing the various concepts and retain
only concepts with at least 10 positive examples (more details in
Section \ref{subsec:Visual-vs-Concept}).

Many of the concepts overlap, such as ``person'' and ``man'',
or are closely related, such as A being a sub-type of B (``cat''
vs ``animal''). Hence it would harm the classifiers to treat them
as mutually exclusive. To overcome this, we represent each concept
by its representation produced by the GloVe method of \cite{pennington2014glove}.
This 300-D representation has been shown to correlate well with semantic
meaning, embedding semantically related words closely together in
vector space, as well as other interesting properties. 

We perform K-means \cite{arthur2007k} clustering on the embedded
word-vectors for a total of 100 clusters. As expected, these produce
semantically meaningful clusters, forming groups such as : (sign,street-sign,traffic-sign,...),(collar,strap,belt,...),

(flower,leaf,rose,...),(beer,coffee,juice), etc. 

We assign to each concept $c$ a cluster $h_{c}$. Denote by $C_{I}$
the set of concepts in an image $I$ according to the ground-truth
in the VG \cite{journals/corr/KrishnaZGJHKCKL16}dataset. We define
the positive and negative training sets as follows:

\begin{eqnarray}
R_{pos}(c) & = & \{I:c\in C_{I}\}\\
R_{neg}(c) & = & \{I:h_{c}\cap C_{I}=\emptyset\}
\end{eqnarray}

In words, $R_{pos}$ is the set of all images containing the target
concept $c$ and $R_{neg}$ is the set of all images which do not
contain any concept in the same cluster as $c$. We sample a random
subset from $R_{neg}$ to avoid highly-imbalanced training sets for
concepts which have few positive examples. In addition, sampling lowers
the chance to encounter images in which $c$ or members of $h_{c}$
were not labeled. In practice, we limit the number of positive samples
of each concept to $1000$ , as using more samples increased the run-times
significantly with no apparent benefits in performance.

The classifiers trained on the set concepts in this way serve as the
$F_{i}$ in Section \ref{subsec:Concepts-as-Classifiers}. We next
proceed to describe experiments validating our approach and comparing
it to standard approaches. 

\section{Experiments}

\begin{figure}
\begin{centering}
\includegraphics[width=.7\columnwidth]{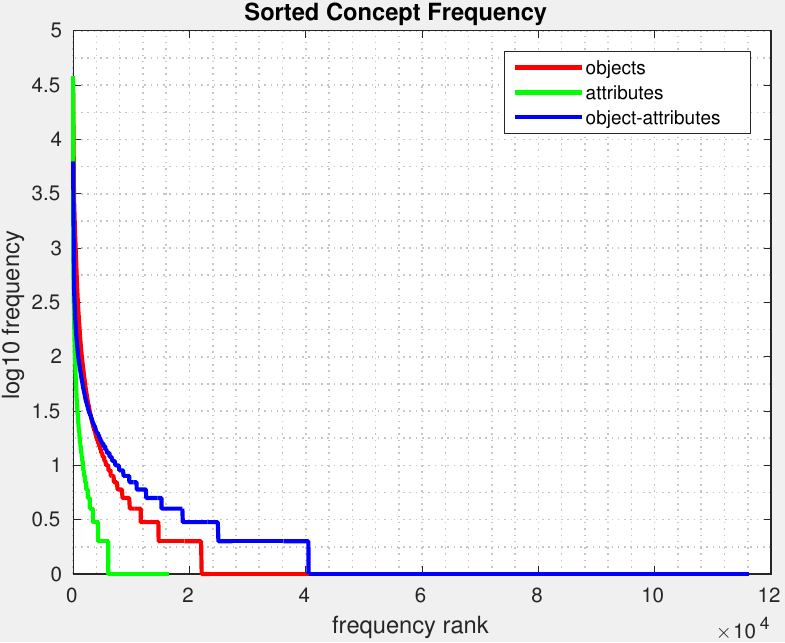}
\par\end{centering}
\caption{\label{fig:(a)-Object-(red)}Object (\textbf{\textcolor{red}{\emph{red}}})
and attributes (\textbf{\textcolor{green}{\emph{green}}}) in the VG
dataset \cite{journals/corr/KrishnaZGJHKCKL16} follow a long tail
distribution. Object paired with attributes (\textbf{\textcolor{blue}{\emph{blue}}})
much more so. }
\end{figure}

To validate our approach, we have tested it on the Standford-40 Action
dataset \cite{yao2011human}. It contains 9532 images with a diverse
set of of 40 different action classes, 4000 images for training and
the rest for testing. In addition we test our method on the recently
introduced HICO \cite{chao2015hico} dataset, with 47774 images and
600 (not all mutually exclusive) action classes. Following are some
technical details, followed by experiments checking various aspects
of the method. 

As a baseline for purely-visual categorization, we train and test
several baseline approaches using feature combinations from various
sources: (1) The global average pooling of the VGG-GAP network \cite{zhou2015learning}
(2) the output of the 2-before last fully connected layer (termed
fc6) from VGG-16 \cite{simonyan2014very} and (3) The pool-5 features
from the penultimate layer of ResNet-151 \cite{he2015deep}. The feature
dimensions are 1024, 4096 and 2048, respectively. In all cases, we
train a linear SVM \cite{journals/jmlr/FanCHWL08} in a one-versus-all
manner on $\ell_{2}$ normalized features, or the $\ell_{2}$ normalized
concatenation of $\ell_{2}$ features in case of using several feature
types. To assign GloVe \cite{pennington2014glove} vectors to object
names or attributes, we use the pre-trained model on the Common-Crawl
(42B) corpus, which contains a vocabulary of 1.9M words. We break
up phrases into their words and assign to them their mean GloVe vector.
We discard a few words which are not found in the corpus at all (e.g.,
``ossicones''). 

\begin{figure}
\begin{centering}

\includegraphics[width=.8\columnwidth]{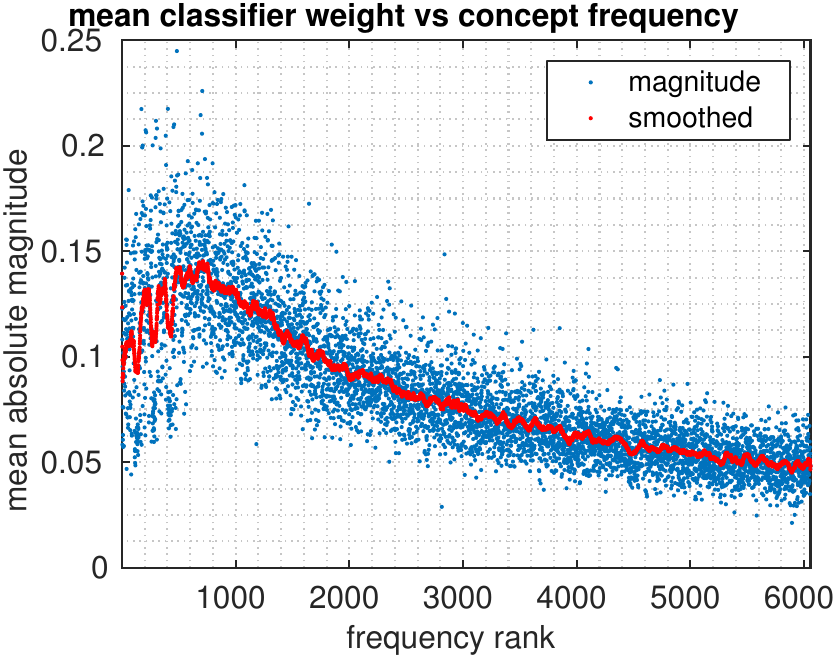}\caption{\label{fig:(blue)-The-magnitude}(\emph{blue}) The magnitude of weights
assigned to different concepts is small for very common (since they
are not discriminative) or very rare concepts (since they are harder
to learn). The smoothed (\emph{red}) dots show the moving average
of the weight with a window size of 50 concepts, to better show the
trend in magnitude. }
\end{centering}
\end{figure}

\subsection{Visual vs Concept features\label{subsec:Visual-vs-Concept}}
\begin{center}
\begin{table}
\begin{centering}
\scalebox{0.9}{
\begin{tabular}{cccc}
\toprule 
 & \multicolumn{3}{c}{Stanford-40\cite{yao2011human}(precision)}\tabularnewline
\midrule 
Method \textbackslash{} Features & G & G+V & G+V+R\tabularnewline
\midrule
\midrule 
Direct & 75.31 & 78.78 & 82.97\tabularnewline
\midrule 
Concept(Obj) & 74.02 & 77.46 & 81.27\tabularnewline
\midrule 
Concept(Attr) & 74.22 & 77.26 & 81.07\tabularnewline
\midrule 
Concept(Obj-Attr) & 38.38 & 33.88 & 34.74\tabularnewline
\midrule 
Concept(Obj)+Direct & 75.31 & 78.81 & \textbf{83.12}\tabularnewline
\midrule 
Other Works & \multicolumn{3}{c}{80.81 \cite{gao2015deep}}\tabularnewline
\bottomrule
\end{tabular}\,%
\begin{tabular}{ccc}
\toprule 
\multicolumn{3}{c}{HICO\cite{chao2015hico}(mAP)}\tabularnewline
\midrule 
G & G+V & G+V+R\tabularnewline
\midrule
\midrule 
24.96 & 28.13 & 31.49\tabularnewline
\midrule 
24.4 & 26.5 & 29.6\tabularnewline
\midrule 
23.9 & 26.12 & 28.85\tabularnewline
\midrule 
38.38 & 33.88 & 34.74\tabularnewline
\midrule 
25.06 & 28.21 & \textbf{31.54}\tabularnewline
\midrule 
\multicolumn{3}{c}{29.4$^{\star}$ / 36.1$^{\dagger}$\cite{mallya2016learning}}\tabularnewline
\bottomrule
\end{tabular}
}
\par\end{centering}
\caption{\label{tab:Classification-using-direct}Classification accuracy/mAp
using direct visual features extracted directly from images (\emph{Direct})
vs. proposed method (\emph{Concept($\cdot$)}) for concatenations
of various feature types. \emph{G}: Global-Average-Pooling layer from
\cite{zhou2015learning}. \emph{V}: fc6 from VGG-16 \cite{simonyan2014very}.\emph{
R: }pool5 (penultimate layer) from ResNet-151\cite{he2015deep}. Describing
images by their set of semantic constituents performs similarly to
learning the direct appearance of the classes. The words in brackets
specify the types of concepts used (objects/attributes/both). Concept(Obj)+Direct:
a weighted combination of the output scores. Rare concepts such as
paired object and attributes perform poorly. $\star$: fine-tuned
the VGG-16 network. $\dagger$: used detected person bounding boxes;
we use the entire image only.}
\end{table}
\par\end{center}

Training and testing by using the visual features is straightforward.
For the concept features, we train concept detectors on both the objects
and object attributes of the VG dataset. Directly using these in their
raw form is infeasible as explained in Sec. \ref{subsec:Refinement-via-Language}.
To reduce the number of classes, we normalize each object name beforehand.
The object name can be either a single word, such as ``dog'', or
a phrase, such as ``a baseball bat''. To normalize the names, we
remove stop-words, such as ``the'',''her'',''is'',''a'', as
well as punctuation marks. We turn plural words into their singular
form. We avoid lemmatizing words since we found that this slightly
hinders performance, for example, the word ``building'' usually
refers to the structure and has a different meaning if turned to ``build''
by lemmatization. Following this process we are left with 66,390 unique
object names. They are distributed unevenly, the most common being
``man'',''sky'',''ground'',''tree'', etc. We remove all objects
with less than 10 images containing them in the dataset, leaving us
with 6,063 total object names. We do the same for object attributes:
we treat attributes regardless of the object type (e.g, ``small dog''
and ``small tree'' are both mapped to ``small''), as the number
of common object-attribute pairs is much smaller than the number of
attributes. Note that the stop-word list for attributes is slightly
different, as ``white'', a common word in the dataset, is not a
noun, but is a proper attribute. We are left with 1740 attributes
appearing at least 10 times in the dataset. See Fig. \ref{fig:(a)-Object-(red)}
for a visualization of the distribution of the object and attribute
frequency. Specifically, one can see that for object-attribute pairs
the long-tail distribution is much more accentuated, making them a
bad candidate for concepts to learn as features for our target task.

We train classifiers using the detected objects, attributes and object-attributes
pairs, as described in sections \ref{subsec:Concepts-as-Classifiers}
and \ref{subsec:Refinement-via-Language}. Please refer to Table \ref{tab:Classification-using-direct}
for a comparison of the direct-visual results to our method. Except
for the attribute-object concepts, we see that the concept based classification
does nearly as well as the direct visual-based method, where the addition
of ResNet-151 \cite{he2015deep} clearly improves results. Combining
the predictions from the direct and concept-based (object) predictions
and using the ResNet features along with the other representations
achieves an improved 83.12\% on Stanford-40 \cite{yao2011human}.
On the recent HICO \cite{chao2015hico} dataset we obtain a mean average
precision of 31.54\%. \cite{mallya2016learning} obtain higher results
(36.1\%) by fusing detected person bounding-boxes with global image
context and using a weighted loss.

\subsection{Describing Actions using Concepts\label{subsec:Describing-Actions-using}}
\begin{center}
\begin{figure}
\begin{centering}
\includegraphics[width=0.35\textwidth]{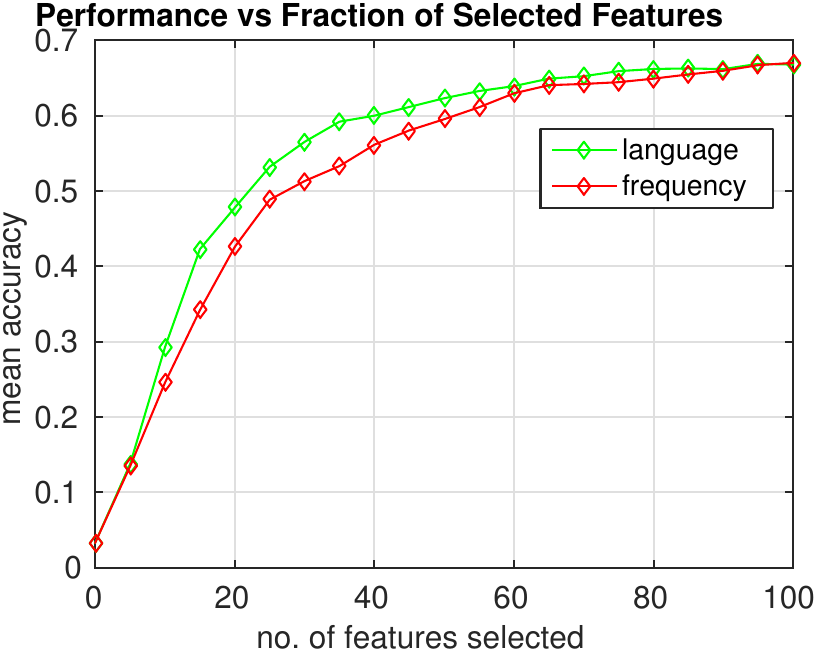}
\par\end{centering}
\caption{\label{fig:Classification-accuracy-by}Classification accuracy by
selecting the first $k$ concepts according to concept frequency (\textbf{\textcolor{red}{\emph{red}}}\emph{)
}vs. semantic relatedness (\textbf{\textcolor{green}{\emph{green}}}).
Semantically related concepts outperform those selected via frequency
when using a small number of features.}
\end{figure}
\par\end{center}

For each target class, the learned classifier assigns a weight for
each of the concepts. Examining this weight vector reveals the concepts
deemed most relevant by the classifier. Recall that the weight vector
of each learned classifier for class $j$ is $\omega_{j}\in\mathbb{R}^{N}$
($N$ is the number of target concepts). We checked if the highest-weighted
concepts carry semantic meaning with respect to the target classes
as follows: for each target class $j\in L$ ($L$ being the set of
target classes in $\mathcal{F}$) we sort the values of the learned
weight-vector $\omega_{j}$ in descending order, and list the concepts
corresponding to the obtained ranking. Table \ref{tab:Highest-ranking-concepts}
shows ten arbitrarily chosen classes from the Stanford-40 Actions
\cite{yao2011human} dataset, with the top 5 ranked object-concepts
according to the respective weight-vector. In most cases, the classes
are semantically meaningful. However, in some classes we see unexpected
concepts, such as holding\_an\_umbrella$\rightarrow$handbag. This
points to a likely bias in the Stanford-40 dataset, such as that many
of the subjects holding umbrellas in the training images also carry
handbags, which was indeed found to be the case by examining the training
images for this class. 

An interesting comparison is the concepts differentiating between
related classes. For example, the top 5 ranked keywords for the class
``feeding a horse'' are (``mane'', ``left\_ear'', ``hay'',
``nostril'', ``horse'') whereas for ``riding a horse'' they
are (``saddle'', ``horse'', ``rider'', ``hoof'', ``jockey'').
While ``horse'' is predictably common to both, other words are indeed
strongly related to one of the classes but not the other, for example,
``hay'' for feeding vs ``jockey'', ``saddle'' for riding. 

\subsection{Concept Visualization}
\begin{center}
\begin{figure}
\begin{centering}
\subfloat[]{\includegraphics[height=0.19\textwidth]{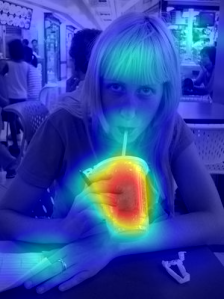}

}\subfloat[]{\includegraphics[height=0.19\textwidth]{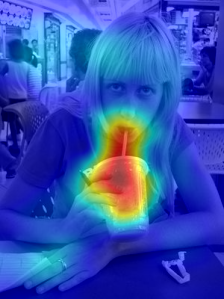}

}\subfloat[]{\includegraphics[height=0.19\textwidth]{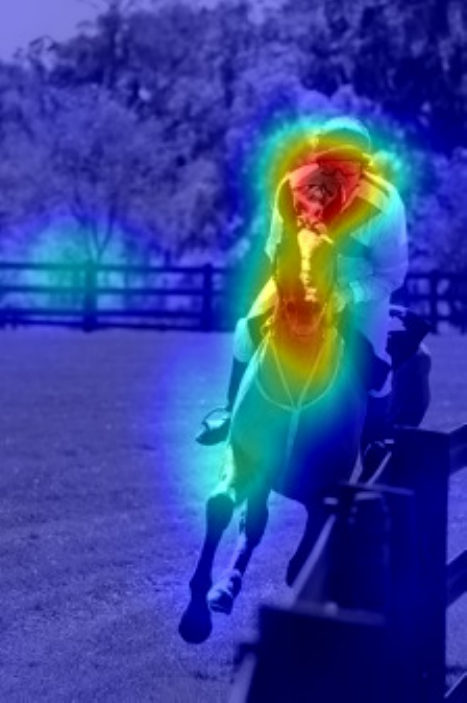}

}
\par\end{centering}
\begin{centering}
\subfloat[]{\includegraphics[height=0.19\textwidth]{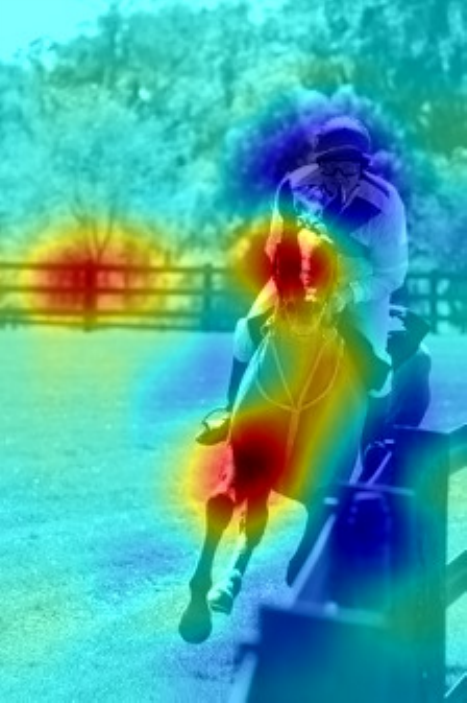}

}\subfloat[]{\includegraphics[height=0.19\textwidth]{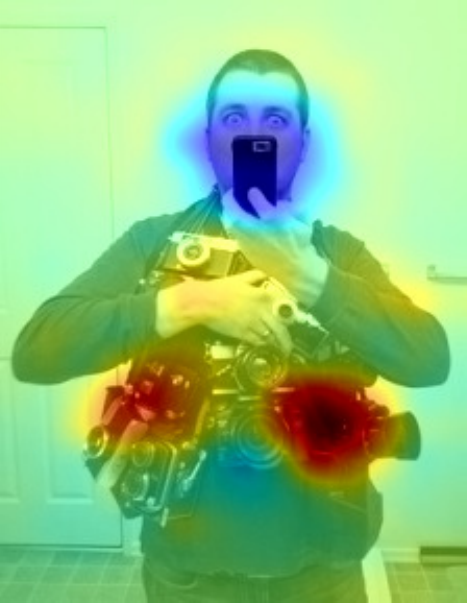}

}\subfloat[]{\includegraphics[height=0.19\textwidth]{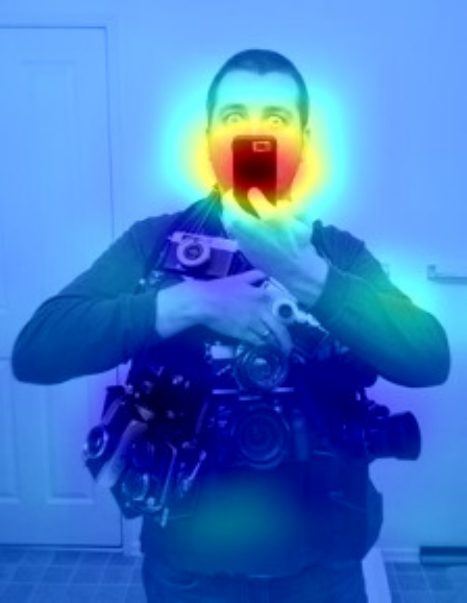}}
\par\end{centering}
\caption{\label{fig:Forcing-the-method}Forcing the method to explain the \emph{wrong
}class: the proposed method detects action classes by a weighted sum
of learned concepts. We visualize (using \cite{zhou2015learning})
highlighted regions contributing to the strongest concepts related
to the correct vs the incorrect class \emph{. }The correct(incorrect)
pairs are : \emph{(a,b)} drinking (smoking) \emph{(c,d)} riding a
horse (feeding a horse) \emph{(e,f)} taking photos (phoning). When
forced to explain the image differently, the method highlights different
concepts, relevant to the desired class: ``smoking'' shifts focus
to the immediate mouth area while ``drinking'' on the cup. ``feeding
a horse'' focuses on the head and lower body of the horse, ``riding
a horse'' on the rider. ``taking photos'' focuses on the cameras,
``phoning'' on the hand-held phone.}
\end{figure}
\par\end{center}

To test what features contribute most to the classification result,
we use the Class-Activation-Map (CAM) \cite{zhou2015learning}.
This method allows to visualize what image regions contributed the
most to the score of each class. We can do this for the version of
our method which uses only the VGG-GAP features, as the method requires a
specific architecture to re-project classification scores to the image (see \cite{zhou2015learning} for details). We visualize the average CAMs of the top-5 ranked keywords for different
classes (as in the above section). We do this for two target classes
for each image, one correct class and the other incorrect, to explain
what image regions drive the method to decide on the image class.
See Fig. \ref{fig:Forcing-the-method}. When the method is ``forced''
to explain the riding image as ``feeding a horse'', we see negative
weights on the rider and strong positive weights on the lower part
of the horse, whereas examining the regions contributing to ``riding
a horse'' gives a high weight to the region containing the jockey. 

\subsection{Distribution of Weights}

We have also examined the overall statistics of the learned weight
vectors. For a single weight vector $\omega$, we define:
\begin{eqnarray}
abs(\omega) & = & [\left\vert \omega_{1}\right\vert ,\left\vert \omega_{2}\right\vert ,\dots,\left\vert \omega_{N}\right\vert ]\\
\bar{\omega} & = & \frac{1}{N}\sum_{j=1}^{L}abs(\omega_{j})
\end{eqnarray}
i.e, $\bar{\omega}$ is the mean of $abs(\omega)$ for all classifiers
of the target classes. Fig. \ref{fig:(blue)-The-magnitude} displays
these mean absolute weights assigned to object concepts, ordered by
their frequency in VG. Not surprisingly, the first few tens of concepts
have low-magnitude weights, as they are too common to be discriminative.
The next few hundreds of concepts exhibit higher weights, and finally,
weights become lower with diminished frequency. This can be explained
due to such concepts having weaker classifiers as they have fewer
positive examples, making them less reliable. A similar trend was
observed when examining attributes.

\subsection{Feature Selection by Semantic Relatedness}

Section \ref{subsec:Describing-Actions-using} provided a qualitative
measure of the keywords found by the proposed method. Here, we take
on a different approach, which is selecting concepts by a relatedness
measure to the target classes, and measuring how well training using
these concepts alone compares with choosing the top-k most common
concepts. To do so, we measure their mean ``importance''. As described
in Section \ref{subsec:Refinement-via-Language} we assign to each
concept $c\in C$ a GloVe \cite{pennington2014glove} representation
$V_{c}$. Similarly, we assign a vector $V_{p}$ to each target class
$p\in L$ according to its name; for instance, ``riding a horse''
is assigned the mean of the vectors of the words ``ride'' and ``horse''.
Then, for each class $p$ we rank the $V_{c}$ vectors according to
their euclidean distance from $V_{p}$ in increasing order. This induces
a per-class order $\sigma_{p}=ex
c_{p,1}\dots c_{p,2}$, which is a permutation
of $1...\left\vert C\right\vert $, such that $c_{p,i}$ is the ranking
of $c_{i}$ in the ordering induced by $p$. We use this to define
the new \emph{mean} rank $r(c)$ of each concept:

\begin{eqnarray}
r(c) & = & {\displaystyle \sum_{p=1}^{L}}\exp(-\sigma_{p}(c))\label{eq:ordering}
\end{eqnarray}

Now, we test the predictive ability of concepts chosen from $C$ according
to two orderings. The first is the frequency of $c$, in ascending
order, and the second is the sorted values (descending) of $r(c)$
as defined in Eqn. \ref{eq:ordering}. We select the first $k$ concepts
for the first $k\in(0,5,10,15,\dots,100)$ features ($k=0$ for chance
performance). For a small amount of features, e.g., $k=15$, the concepts
chosen according to $r(c)$ outperform those chosen according to frequency
by a large margin, i.e, 42.2 vs 34.2 respectively. 

\section{Conclusions \& Future Work}

We have presented a method which learns to recognize actions in images
by describing them as a weighted sum of detected concepts (objects
and object attributes). The method utilizes the annotations in the
VG dataset to learn a broad range of concepts, which are then used
to recognize action in still images. We are able to improve on classification
performance versus of a strong baseline which uses purely visual features,
as well as provide a visual and semantic explanation of the classifier's
decisions. In the future we intend to broaden our work to capture
object relationships, which are very important to action-classification
as well.

\small\bibliographystyle{plain}
\bibliography{bibtex}

\end{document}